# WORD PLAY FOR PLAYING OTHELLO (REVERSES)

Samantha E. Miller Noever, David Noever[1]

[1]PeopleTec, Inc., Huntsville, AL, USA
david.noever@peopletec.com

**ABSTRACT**
Language models like OpenAI's Generative Pre-Trained Transformers (GPT-2/3) capture the long-term correlations needed to generate text in a variety of domains (such as language translators) and recently in gameplay (chess, Go, and checkers). The present research applies both the larger (GPT-3) and smaller (GPT-2) language models to explore the complex strategies for the game of Othello (or Reverses). Given the game rules for rapid reversals of fortune, the language model not only represents a candidate predictor of the next move based on previous game moves but also avoids sparse rewards in gameplay. The language model automatically captures or emulates championship-level strategies. The fine-tuned GPT-2 model generates Othello games ranging from 13-71% completion, while the larger GPT-3 model reaches 41% of a complete game. Like previous work with chess and Go, these language models offer a novel way to generate plausible game archives, particularly for comparing opening moves across a larger sample than humanly possible to explore. A primary contribution of these models magnifies (by two-fold) the previous record for player archives (120,000 human games over 45 years from 1977-2022), thus supplying the research community with more diverse and original strategies for sampling with other reinforcement learning techniques.

## INTRODUCTION

Human-played games fascinate machine learning researchers as challenging testbeds and benchmarks for artificial intelligence (AI) (Goecks et al. 2022). A 2020 review of AI growth focuses on the critical decades (1993-2012) as a zenith in brute-force search and rule-based methods (Moy et al., 2020). During this period, researchers often combined heuristics tailored by human experts with brute-force enumeration of all available and plausible moves to defeat world champions in chess, checkers, and Othello (Cherry, 2011).

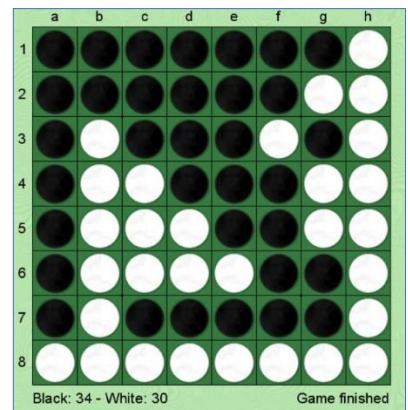

Figure 1. Othello game replayed from World Championship archive (Hiroshi vs. Thomas, 1977).

*Games in AI.* For AI researchers, game scores provide simple reward and penalty functions to explore algorithmic search methods and rank machine learning progress (Schrittwieser et al., 2020). Competitive games naturally offer a score (reward) and an archive of human players' history to learn from or compare win-loss records as novel approaches to rank order the best-emerging algorithms (Kasparov, 2018). Stages of AI progression mainly appear when the game and algorithms combine to include direct play between humans versus machines (Cherry, 2011). Game designers often follow a simplified formula for maximum player engagement if the rules are quick to pick up but hard to master. Good games have a combinatorial expansion of possible moves, making simple move restrictions translate into deep strategic complexity over many years of gameplay. Humans learn the strategy of a well-designed game in five minutes or less (e.g., Go), while mastery requires a lifetime of study and experience. Most two-player games typically tax human patience if not completed in hours rather than days. This paper addresses one novel approach to explore Othello (Figure 1), principally whether a language model not designed for this task (Ciolino et al., 2020; Noever et al., 2020) can offer a plausible opponent in a challenge with often unclear reward structures and quick reversals of fortune.

According to *Games of Many Nations* (Harbin, 1954), the game of Othello originated in China, but other historical accounts place the game in either 1880's England or post-war Japan (Othello, 2022). Other countries adopted Othello by different names: Reverses, Fan Mien, Reversi, Ping Guo Qi, Apple Chess, Hei Bai Qi, and Black and White Chess (Rose, 2019). Othello's popularity claims 40 million distributed boards sold across 100 countries (Othello, 2022; World Othello Federation, 2022). As a two-player game, the game board resembles chess and checkers (8x8 grid) in 64 identical squares and alternating colors arranged in a checkerboard pattern (Figure 1). Othello resembles a checkers game without removing captured pieces but turning them to future advantage in a territorial grab. Each player starts with two center disks and expands their territory by capturing their opponent's pieces after bracketing or outflanking a line or diagonal to flip their color (black or white). The descriptive name, Reverses, describes the goal of converting an opponent's pieces when sandwiched between two pieces of one's color. Like Go, the Reverses strategy hinges on capturing battlefield territory and planning for strategic or safe positions like corners or walls that thwart an opponent. Like Sudoku games (Ashlesh et al., 2020) that serve as therapy puzzles for cognitive stimulation, notable medical uses of Othello include rehabilitation for patients with acquired brain injuries like stroke and encephalopathy (Kikuchi, 2016).

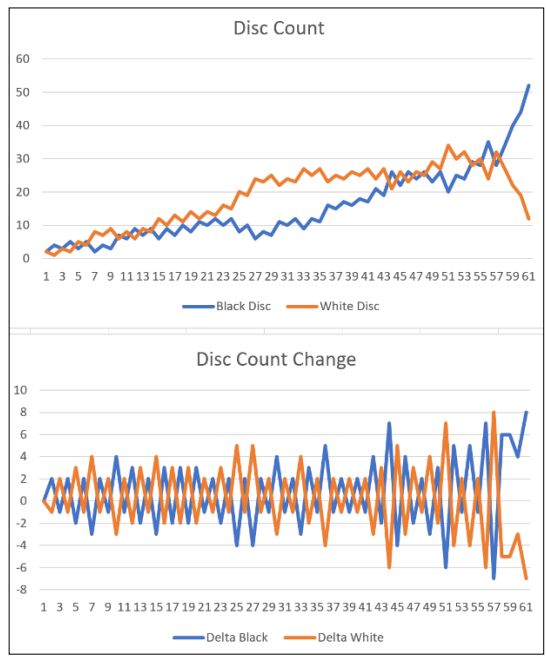

Figure 2. Example Scoring Reversal Timeline in Lopsided 52-12 Black Victory in World Othello Championship, (Thomas vs. Carol, 1977)

*Research Motivations.* The difficulty in mastering Othello is that the game is straightforward to learn but challenging to master (Glessner, 2020). The winning player claims the most pieces of their color on the board at the game's end. However, Othello can be very complex, with many strategies and tactics that reverse fortunes (Noever and Burdick, 2021). Othello offers some unique opportunities for machine learning challenges relative to other board games. First, in Othello, it is much easier for a player to make a mistake and lose control of the disc count (Marcin, 2017). In other games like chess, it is much easier to recover from errors. Because of stair-stepping rewards (Figure 2), a diagonal or corner position in Othello can render an opponent with few options. In the example shown selected from the lopsided 52-12 World Championship game, the winning player (Black) leads the losing player (White) for 15 turns out of 60 moves. So despite the 40-piece final margin of victory, the winning side lost 75% of the turns (Pederson, 2022; Othello, 2022). Similarly, a single mistake in outflanking an opponent can give an opponent an unconquerable advantage. Second, the pieces in Othello have one flanking function, but only relative to an opponent's position, making the game seem more like a puzzle than a game (Tieperman, 2020). Compared to chess, where the pieces have a more general function of moving on exclusive diagonal or linear paths, Othello's rules naturally constrain players to consider their current position alongside any anticipated adversarial responses. Third, the Othello board is symmetrical to both players, while the chess board offers a definite attacking or defending side. In other words, chess and checkers present radically different views from the other side of the gameboard compared to Othello. This Othello symmetry can make it challenging to see reverse positional patterns and an adversary's plan. Othello requires strategic thinking and planning, even though the moves are straightforward and follow the maxim "to support full game explanation within five minutes but to offer sufficient strategic challenge to interest expert game study."

*Previous Othello Successes.* The challenge of algorithmic solutions for Othello stems partly from the large search space along with the rugged reward and penalty functions (Figure 2). Lee (1990) and later Buro (2003) presented the computational requirements for strong-playing Othello programs. Cherry (2011) demonstrated that an intelligent Othello player should combine machine learning with game-specific heuristics. The human-machine combination echoes broad aspects of fine-tuning a language model like GPT-2 using previous human (text) archives for Othello. Eskin et al. (1999) addressed finding Othello moves as a complex search problem that was amenable to evolutionary strategies and genetic programming between available "chromosomes" of effective moves. Hingston and Masek (2007) experimented with a distribution of trial-error solutions and Monte Carlo simulations to narrow the search space for the next move generation. Like other board games such as chess and checkers, the machine mastery of strategy has long surpassed human expert limits and defeated grandmasters. Earlier attempts in machine learning to play Othello include using artificial neural networks and Monte Carlo tree search, either by playing machine vs. machine training rounds from scratch or by augmenting the performance of a human player based on heuristics and archived gameplay. Van Eck and van Wezel (2008) and Van Der Ree and Wiering (2013) applied reinforcement learning to play Othello. Although not explicitly included in the AlphaZero list of games, a recent application to 6x6 Othello (Chang et al., 2018; Thakoor et al., 2017) validated the Monte Carlo tree search application to restrict the search for the next moves. Some examples of using the Monte Carlo tree search to devise strategies to play Othello are as follows: 1) explore the space of possible moves, and choose the move that results in the most favorable outcomes; 2) evaluate the potential outcomes of each move, and choose the move that is most likely to lead to a winning position, and 3) generate a large number of random game positions, and use these to estimate the probabilities of each move leading to a win. Ultimately, the reinforcement learning method employs a more intelligent trial-and-error agent whose only narrow AI task is to choose the move with the highest probability of leading to a win. Liskowski et al. (2018) showed the first deep learning approach to learning Othello with neural networks. In other games like chess (Toshniwal et al., 2020; Swingle, 2017) and Go (Ciolino et al., 2020), the transformer model with attention (Vaswani et al., 2017) yields attractive alternatives for exploring the game search space. To the authors' knowledge, little previous work has applied language modeling to solve Othello's next move generation. The lack of easily transformed game archives (in Portable Game Notation or PGN, Pederson, 2022)) may have delayed implementations compared to Go or chess language modeling (Figure 3).

**METHODS**

Unlike previous neural network methods, transformers learn language representations from scratch without pre-training (Brown et al., 2020). Such initialization or "few-shot" learning is made possible by using attention mechanisms, which allow the transformer to focus on specific parts of the input sentence at each step of its transformation process. By over-weighting likely language pairs, transformer models maintain longer-term correlations needed to master text generation tasks. This long-range word model results in a much more efficient and effective learning process and allows the

```
1. F5 D6 2. C3 D3 3. C4 F4 4. F6 G5 5. E6 F7
6. C7 C5 7. G3 B5 8. E3 G4 9. B3 C6 10. D7
E7 11. B6 B4 12. F3 D8 13. A6 E2 14. F1 C8
15. A5 H3 16. F2 D2 17. C2 B2 18. G6 H5
19. A1 D1 20. E1 G1 21. H7 A3 22. A2 A4
23. B1 H6 24. G7 H8 25. E8 F8 26. H4 B7
27. B8 C1 28. H1 A8 29. A7 G8 30. H2 G2
```

Figure 3. 30 Othello Turns in Completed World Championship (1977) Encoded as ASCII Text

transformer to achieve state-of-the-art results on various tasks. An intriguing game combination arises if the language model can simulate novel game output and subsequently play in a visualized game environment as an actual human or machine player might engage an opponent (Othbase, 2022). This visualization approach has previously offered a concrete method for visualizing abstract language instructions (or parroted game notations) but as if responding in a contextually-rich strategy to current game situations.

*GPT-2 as a fine-tuned language model*. GPT-3 represents a milestone in the current state-of-the-art for accessible language models (Floridi and Chirliatti, 2020). However, the much smaller GPT-2 models can produce expressive output that exceeds the GPT-3 paradigm, which fundamentally seeks to be a zero-shot or no-preparation language model (Woolf, 2019). One can, for instance, ask GPT-3 for sentence (or prompt) completion that is not ideally suited for its training on the internet-wide (Reddit) dataset. A favorite challenge might be to write an essay in the style of Hemingway or Shakespeare (Dale, 2021). Unlike GPT-2, which requires more than 1000 training sentences, GPT-3 mimics writer style given a single sentence prompt (OpenAI, 2022). Despite this immediate grasp of GPT-3 for contextual cues, GPT-2 remains a smaller but accessible approach to specialized domains (Klein and Nabi, 2019). The GPT-2 methods employed in the current research include using a domain-specific corpus (Woolf, 2019). Figure 4 illustrates a typical learning curve associated with acquiring domain-specific expertise from plain language inputs like Othello game archives. This type of corpus is typically smaller than the original GPT-2 corpus or the GPT-3 internet sample but contains more relevant data to the task at hand. Fine-tuning helps the transformer learn domain-specific vocabulary and syntax. Secondly, fine-tuning GPT-2 can uniquely offer player-sidedness using a labeled dataset. We have previously modeled chess opening moves (black only) by presenting a sided point of view of the game history for fine-tuning (Noever et al., 2020; Ciolino et al., 2020; Noever and Burdick, 2021). Similarly, one can explore asymmetric game odds by fine-tuning text archives that favor one style of play (e.g., grandmaster ELO rank) or even one preferred opening move. Human archives of previous games can train the transformer to perform other specific tasks, such as classification or question answering (Pederson, 2022; Othello;

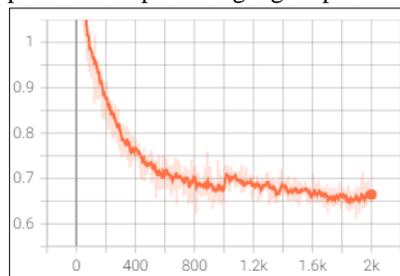

Figure 4. Loss rate from Othello training data over 2000 epochs

2022). Finally, fine-tuning accelerates training performance in a transfer learning approach. This fine-tuned approach allows the transformer to learn from a pre-trained or partial model, such as the original GPT-2 transformer trained on much larger language or sentence completion starting datasets (Kumar et al., 2020). Transfer learning can help the transformer learn faster and improve performance on the task (Zhuang et al., 2020). In addition to fine-tuning, other design methods (Narang et al., 2021) for transformers can overcome previous AI shortcomings, including 1) Increasing the number of hidden layers in the transformer. 2) Adding more neurons to the hidden layers. 3) Changing the activation function used in the hidden layers. 4) Using a different optimizer for training the transformer. 5) Adding regularization techniques to the transformer.

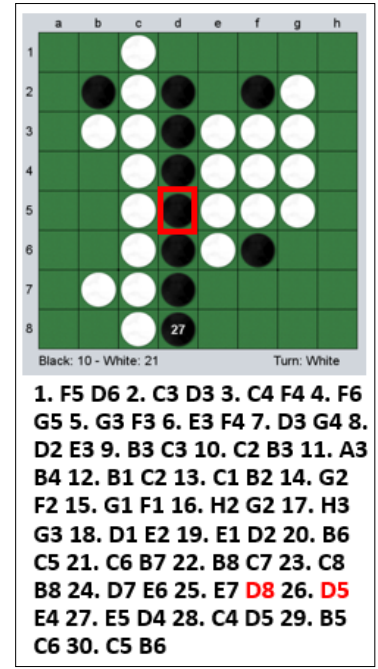

*Othello archives to fine-tune language models.* The research thesis values a visualization component unique to game simulation and harvesting from human knowledge in the format of game archives with actual win-loss results (Othbase, 2022). The favored (ASCII) text format for simplicity is Portable Game Notation (PGN) which dominates board games like chess and checkers (Pederson, 2022; Brown et al., 2017). While Othello shares the same 8x8 gameboard, it employs its archival format, which is convertible to plain text records that showcase the x-axis (A-H) and y-axis (1-8) positions alternating with the black first and white second. As shown in Figure 5, a complete Othello game (starting from 4 initial pieces) thus describes 30 subsequent turns for each player. Several key features of PGN archives make them ideal for training language models to play games (Pederson, 2022). First, PGN archives are typically organized by play so that each game is a self-contained unit. Records of winning moves make it easy for the language model to learn the game's rules from the PGN archive. Second, PGN archives often include extensive commentary or metadata on the game history. The metadata can help the language model understand the strategic concepts involved in Othello, particularly for scoring victorious strategies. Third, PGN archives often include multiple variations of each game so that the language model can learn from different types of game styles and player strategies. Finally, PGN archives are usually available in multiple languages, so the language model can learn from various sources. Othello has a strong multi-cultural element in the 100 countries that play (and archive) game histories.

Figure 5. Early termination of game play at step 26 of 30 when GPT-3 issues an illegal move

*Fine-tuned GPT-2 plays Othello from archived human games.* To fine-tune GPT-2 requires several thousand text examples (Woolf, 2019). After successively presenting text game play to the pre-trained transformer, the new style of the text supports narrative outputs that mirror the training data. For training Othello, the PGN archive consisted of 125,315 games written with a descriptive header followed by thirty alternative moves (40 megabytes). The maximum tokens (including blank spaces) are around 180 characters for actual game moves, with black assumed to move first in all cases. All recorded games are championship level and include 30 steps with black and white alternating turns. Following the fine-tuning recommendations (Woolf, 2019), we sub-sampled randomly 20,000 games (6 MB), which train GPT-2 on 3.9 million tokens or typically one black-white cycle (4 characters). We used the small GPT-2 model (124 million parameters), which takes approximately 0.5 GB when stored on a disk. The fine-tuning procedure uses 1000-2000 epochs (time steps) with a relatively slow learning rate ($1 \times 10^{-5}$) to preserve the language features of the pre-trained model. We train with a high temperature (originality) parameter, enhancing text generation diversity for future gameplay. The standard length of a text line is limited to 188, which includes a prefix and suffix delimiter ("<|startoftext|>…<|endoftext|>). These delimiters clip the text generator to natural game lengths without extending beyond the 30 alternative steps. After 1000 training steps, the learning plateaus with a minimum and average loss value (0.69 and 0.75, respectively). Figure 6 shows the primary goal of the work in graphical formats. More than 70% of the Othello moves for both players can be emulated by a language model trained on human player history (120,000 previous games). Given the simulation endpoints, an augmented human player can skip the illegal moves that terminate the simulation and continue to play as both or one machine color of discs (black or white). The result is noteworthy (Tsang, 2019) on an 8x8 grid since if one programmed a naïve Othello player, exploring all possible moves represents an example of a combinatorial explosion. If the language model explores up to k moves, the problem challenge increases exponentially (with k moves). From the first move for black discs, there are four possible first moves (f5,e6,c4,d3), and those options are without planning for the opponent's counter moves. In chess, for instance, a grandmaster player may try to look ten moves ahead, which computationally considers trillions of options (Tsang, 2019).

*GPT-3 as an alternative language model.* The larger GPT-3 language model offers a few-shot alternative without relying on explicit fine-tuning of game moves. Instead, the online model (OpenAI, 2022) employs a user prompt to set the initial context and completes a game from the opening moves. For example, if prompted, the transformer answers with an extended Othello game without any other input training: "*You are an expert Othello player. If you are playing black discs, what is your best game from this starting point to a finished 30-move game? The game starts with the following in Portable Game Notation or PGN: 1. F5 D6 2. C3 D3 3. C4 F4 4. F6 G5*". The model responds with a text version of a completed game. The GPT-3 model can generate Othello moves without any previously defined examples.

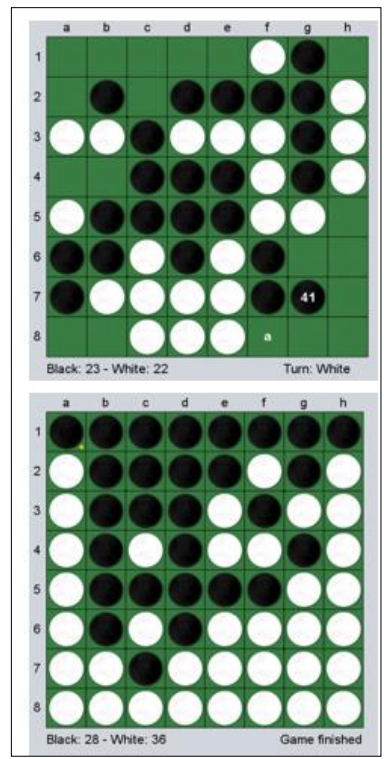

Figure 6. Autogenerated Othello game (top) and completed human gameboard (bottom)

The model reacts correctly to the instructions, terminating the black-white move sequence after 30. Without pre-training, an unexpected prompt like "play to 60 moves instead of 30" will generate an unrealistic PGN file that violates the basic rules. Two illegal moves are noteworthy: 1) placing a disc on an occupied position (Figure 5) and 2) placing an orphaned disc without flanking an opponent or sandwiching enemy positions. The billions of model weights obscure the origin of these game-ending moves. The long-term correlations between disc positions are considerable throughout even the 180-character string representing a completed game. Since the language model generates the next character in a chain (word or sentence completion), the spatial aspects of disc positions decay rapidly, even in large models. Although remarkable in its ability to grasp the question, the answer shows a fundamental lack of understanding of the game rules. While fine-tuned GPT-2 suffers similar deficits in understanding game rules, the strategic use of text delimiters eliminates the user intervention to truncate unphysical extended games.

**RESULTS**

Figure 4 illustrates the rate of learning text archives for Othello play. After 2000 training steps, the loss rate plateaus, and the model ceases learning additional helpful information from a 20,000 sample game input. The longest completed game using GPT-3 without access to the game archive halted with an illegal move after completing 45% of an Othello game (27/60). The longest completed GPT-2 game benefited from access to the library of human play and completed 71% of an Othello game (43/60). When the Othbase (2022) game player imports and displays these generated games, play ceases when a generated game violates the rules. A human player, augmented by the remainder of the generated text game, can complete the game successfully and log a valid score between black and white. Once the customized (Othello-playing) GPT-2 model completes its training cycle, the resulting model generates new games. The GPT-2 model exceeded the number of previous games compiled in a single Othello repository (120,000 human games, Pederson, 2022; Othello, 2022). Exceeding the recorded human player history with a tunable diversity of original moves ("temperature") offers alternative ways to increase the training data and augment the heuristics that human Othello players previously employed over the last 45 years (1977-2022).

Figures 7-8 show that the language transformer generates Othello moves that follow recommended strategies, such as not leaving isolated discs or wedges, particularly next to an open corner. The gameboards represent plausible Othello tactics even when not completed. Since the model develops moves for both players (black and white), the alternating play evolves following the

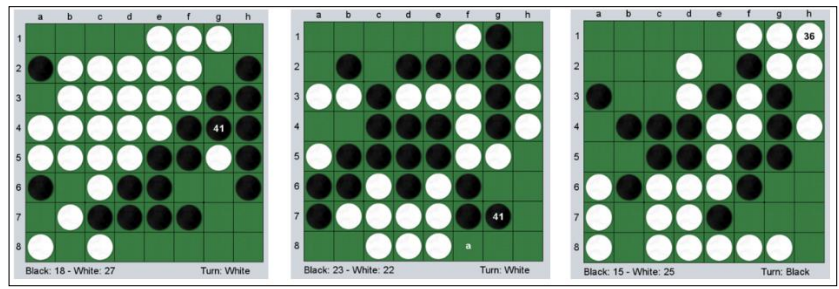

Figure 7. Diversity of opening gameplay using machine-only text generators

championship history, but neither side claims a superior strategy after many trials. Since PGN encodes the outcome, further experiments could easily portray a lopsided game. A language model fine-tuned on dominant black victories would generate more of the same victorious black winning outputs given the filtered inputs. When the developed games play in rapid animation, the language models show an expected diversity of horizontal, vertical, and diagonal captures. The model's scores appropriately acknowledge the expected disc count for each side.

**CONCLUSIONS**

The principal research result of applying language models to games involves demonstrating that the same long-term correlations needed to understand a sentence can alternately learn a complex game move. Othello's strategy offers a challenging test given the rapid reversal of fortunes (Tieperman, 2020; Marcin, 2017). Unlike sparse reward games (Noever and Burdick, 2021), Othello's rules enable not just a rapid change of control or rewards but similarly a steep loss or penalty (Figure 2). As the transformer learns from previous human games (GPT-2) or proposes few-shot alternatives (GPT-3), their results illustrate mapping games' interpretive and contextual power into a surrogate language. We demonstrate that few-shot learning can complete 45% of a game without human supervision simply by understanding the question "to play" or continuing a custom prompt. We also demonstrate that a fine-tuned alternative language model (GPT-2) can specialize its text generation to Othello play exclusively, independent of any user prompt or inputs. Learning from 20,000 championship Othello archives (human players), the GPT-2 model generates 120,000 new Othello games. The text generator completes a single game further than GPT-3's general model but falls short of closing a full 60-move game. The most extended machine-only simulation achieves 71% of an original game before terminating prematurely because of an illegal move (Figure 7). One byproduct of coupling a text archive to a game simulator (like Othbase) enables complex visualization of strategy and score tracking as the algorithm moves to its end state. Future work can extend this powerful combination of language transformers and visualized gameplay to model machine-machine interactions and discover novel strategies for Othello, including self-commentary on expected move scoring.

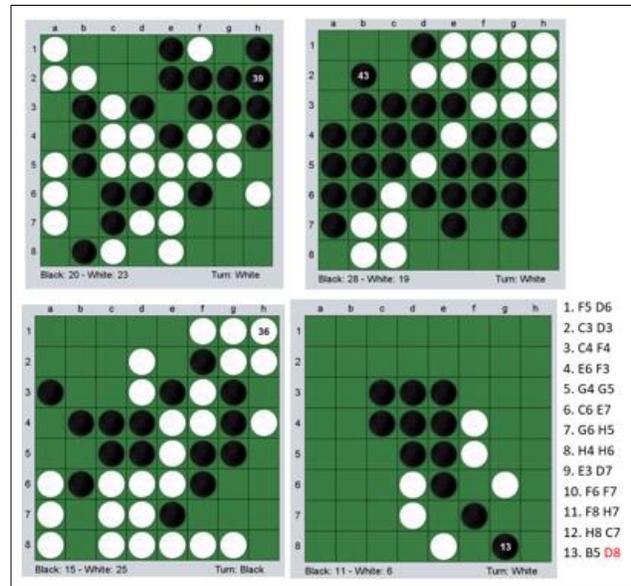

Figure 8. Three long machine-generated games and one short comparison with an illegal move

**ACKNOWLEDGEMENTS**

The authors would like to thank the PeopleTec Technical Fellows program for its encouragement and project assistance.